\pdfoutput=1

\documentclass[11pt]{article}

\usepackage[]{acl}

\usepackage{times}
\usepackage{latexsym}

\usepackage[T1]{fontenc}

\usepackage[utf8]{inputenc}

\usepackage{microtype}
\usepackage{footnotehyper}
\usepackage{tabularx, booktabs}
\usepackage{amssymb}
\usepackage{pifont}
\usepackage{graphicx}
\newcommand{\cmark}{\ding{51}}%
\newcommand{\xmark}{\ding{55}}%
\definecolor{Mycolor1}{HTML}{B5739D}
\definecolor{Mycolor2}{HTML}{67AB9F}
\definecolor{Mycolor3}{HTML}{085EBD}

\title{"Do you follow me?": \\ A Survey of Recent Approaches in Dialogue State Tracking}

\author{Léo Jacqmin\textsuperscript{1,2} \quad Lina M. Rojas-Barahona\textsuperscript{1} \quad Benoit Favre\textsuperscript{2} \\
   \textsuperscript{1}Orange Innovation, Lannion, France \\ 
    \textsuperscript{2}Aix-Marseille Université / CNRS / LIS, Marseille, France \\ 
    \texttt{\{leo.jacqmin, linamaria.rojasbarahona\}@orange.com}  \\
    \texttt{benoit.favre@lis-lab.fr} 
    }
\begin{document}
\maketitle

\begin{abstract}
While communicating with a user, a task-oriented dialogue system has to track the user's needs at each turn according to the conversation history. This process called dialogue state tracking (DST) is crucial because it directly informs the downstream dialogue policy. 
DST has received a lot of interest in recent years with the text-to-text paradigm emerging as the favored approach.
In this review paper, we first present the task and its associated datasets.
Then, considering a large number of recent publications, we identify highlights and advances of research in 2021-2022. 
Although neural approaches have enabled significant progress, 
we argue that some critical aspects of dialogue systems such as generalizability are still underexplored. 
To motivate future studies, we propose several research avenues. 
\end{abstract}

\section{Introduction}
Since human conversation is inherently complex and ambiguous, creating an open-domain conversational agent capable of performing arbitrary tasks is an open problem. Therefore, the practice has focused on building task-oriented dialogue (TOD) systems limited to specific domains such as flight booking. These systems are typically implemented through a modular architecture that provides more control and allows interaction with a database, desirable features for a commercial application.
Among other components, dialogue state tracking (DST) seeks to update a representation of the user's needs at each turn, taking into account the dialogue history. 
It is a key component of the system: the downstream dialogue policy uses this representation to predict the next action to be performed (e.g. asking for clarification). A response is then generated based on this action. 
Beyond processing isolated turns, DST must be able to accurately accumulate information during a conversation and adjust its prediction according to the observations to provide a summary of the dialogue so far. 

\begin{figure}[h]
\centering
\includegraphics[width=1\linewidth]{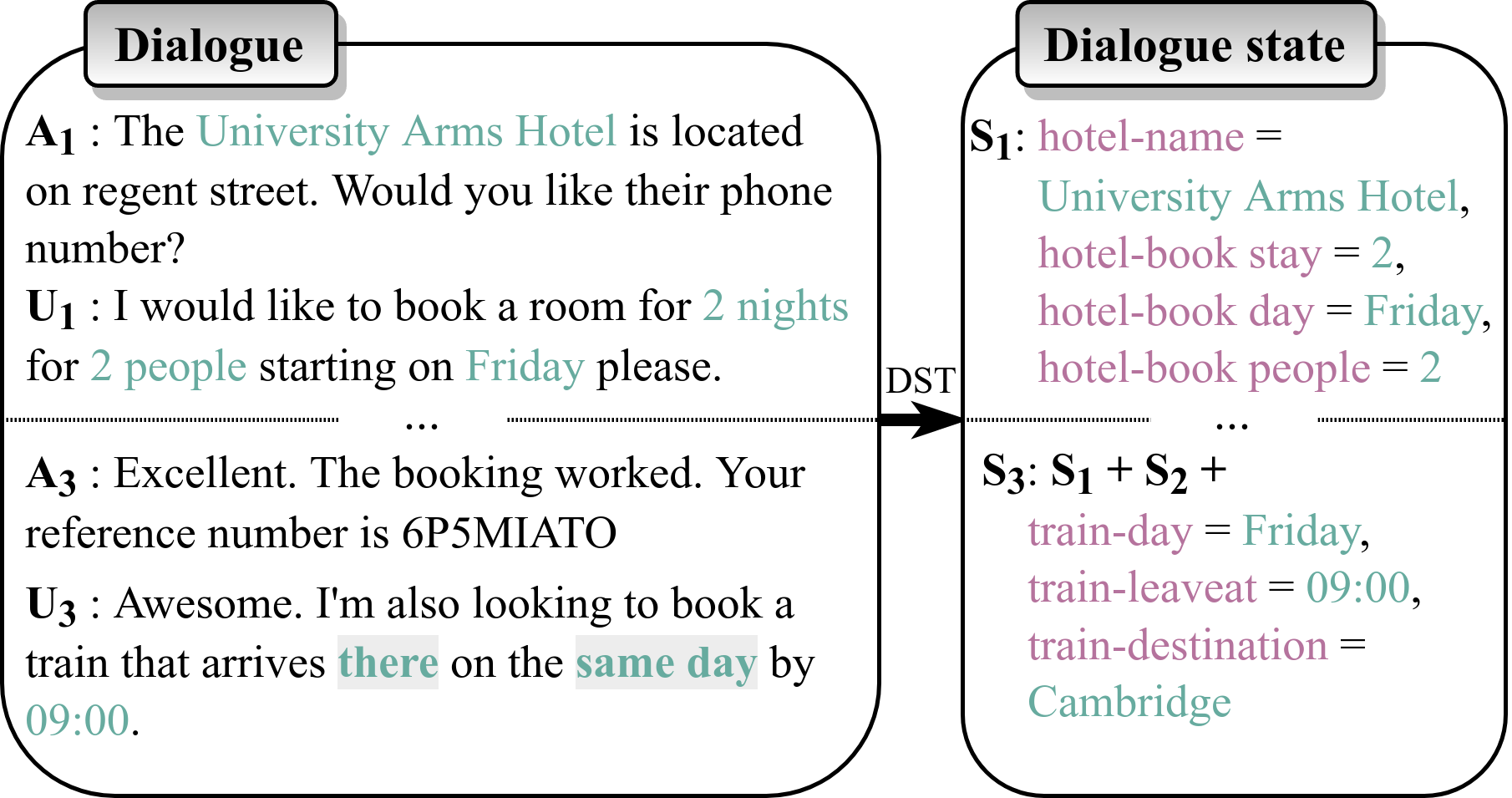}
\caption{Dialogue state tracking (DST) example taken from MultiWOZ. DST seeks to update the dialogue state at each turn as (\textcolor{Mycolor1}{slot}, \textcolor{Mycolor2}{value}) pairs. Current DST models struggle to parse $U_3$, which requires world knowledge and long-term context awareness.}
\label{fig:dst}
\end{figure}

Several papers have surveyed DST from the perspective of a specific paradigm or period.
\citet{youngPOMDPBasedStatisticalSpoken2013a} give an overview of POMDP-
based statistical dialogue systems.
These generative approaches model the dialogue in the form of a dynamic bayesian network to track a belief state. They made it possible to account for the uncertainty of the input related to speech recognition errors and offered an alternative to conventional deterministic systems
which were expensive to implement and often brittle in operation. 
However, POMDP-based systems presuppose certain independencies to be tractable and were gradually abandoned in favor of discriminative approaches that directly model the dialogue state distribution. 
These methods, surveyed by \citet{hendersonMachineLearningDialog2015}, rely on annotated dialogue corpora such as those from the Dialogue State Tracking Challenge, the first standardized benchmark for this task \cite{williamsDialogStateTracking2016}. 
The past years have been marked by the use of neural models that have allowed significant advances, covered by \citet{balaramanRecentNeuralMethods2021}.
Since then, many works dealing with key issues such as adaptability have been published. The first contribution of this paper is a synthesis of recent advances in order to identify the major achievements in the field.

The recent development of neural models has addressed fundamental issues in dialogue systems. 
DST models can now be decoupled from a given domain and shared across similar domains \cite{wuTransferablemultidomainState2019}. They even achieve promising results in zero-shot scenarios \cite{linZeroShotDialogueState2021}.
Despite these advances, modern approaches are still limited to a specific scenario as dialogues in most DST datasets consist of filling in a form: the system asks for constraints until it can query a database and returns the results to the user. 
Moreover, these data-driven approaches are rigid and do not correspond to the need for fine-grained control of conversational agents in an application context. In a real-world scenario, access to annotated data is limited but recent DST models appear to have poor generalization capacities \cite{liCoCoControllableCounterfactuals2021}.
These limitations show that there remain major challenges to the development of versatile conversational agents that can be adopted by the public. 
The second contribution of this paper is to propose several research directions to address these challenges.

\section{Dialogue State Tracking} \label{sec:dst}




A successful TOD system makes it possible to automate the execution of a given task in interaction with a user. The last few years have seen an increase in research in this field, which goes hand in hand with the growing interest of companies in implementing solutions based on TOD systems to reduce their customer support costs.

A typical modular architecture for these systems is composed of the following components: natural language understanding (NLU), DST, dialogue policy, and natural language generation (NLG).
NLU consists of two main subtasks, namely, intent detection 
which consists in identifying the user's intent, such as booking a hotel, 
and slot filling
which consists in identifying relevant semantic concepts, such as price and location.
DST aims to update the dialogue state, a representation of the user's needs expressed during the conversation.
The dialogue policy predicts the system's next action based on the current state. Along with DST, it forms the dialogue manager which interfaces with the ontology, a structured representation of the back-end database containing the information needed to perform the task. 
Finally, NLG converts the system action into natural language. 

In the case of a spoken dialogue system, automatic speech recognition (ASR) and speech synthesis components are integrated to move from speech to text and back. NLU then operates on the ASR hypotheses and is denoted SLU for spoken language understanding.\footnote{Another possibility is end-to-end approaches which seek to obtain a semantic representation directly from speech.}
Traditionally, DST operates on the output of SLU to update the dialogue state by processing noise from ASR and SLU. 
For example, SLU may provide a list of possible semantic representations based on the top ASR hypotheses. DST manages all these uncertainties to update the dialogue state. 
However, recent DST datasets are collected in text format with no consideration for noisy speech inputs, which has caused slot filling and DST to be studied separately. 

Note that such modular architecture is one possible approach to conversational AI and others have been considered. Another approach is end-to-end systems which generate a response from the user's utterance using a continuous representation.
Such systems have been studied extensively for open-domain dialogue \cite{huangChallengesBuildingIntelligent2020} and have also been applied to TOD  \cite{bordesLEARNINGENDTOENDGOALORIENTED2017a}.
 

\subsection{Dialogue State} 


Consider the example dialogue shown in Figure \ref{fig:dst}. Here, the agent acts as a travel clerk and interacts with the user to plan a trip according to the user's preferences.
The dialogue state is a formal representation of these constraints and consists of a set of predefined slots, each of which can take a possible value, e.g. \texttt{book day} = \texttt{Friday}.
At a given turn $t$, the goal of DST is to extract values from the dialogue context and accumulate information from previous turns as the dialogue state $S_t$, which is then used to guide the system in choosing its next action towards accomplishing the task, e.g. finding a suitable hotel. 
Traditionally, informable slots (constraints provided by the user, e.g. \texttt{price range}) are distinguished from requestable slots (information that the user can request, e.g. \texttt{phone number}). 
Informable slots can take a special value: \texttt{don't care} when the user has no preference and \texttt{none} when the user has not specified a goal for the slot.
The current dialogue state can be predicted either from the previous turn using the previous dialogue state $S_{t-1}$ as context, or from the entire dialogue history, predicting a new dialogue state $S_t$ at each turn.

The dialogue state representation as (\textit{slot, value}) pairs is used to deal with a single domain. When dealing with a multidomain corpus, this approach is usually extended by combining domains and slots to extract (\textit{domain-slot}, \textit{value}) pairs.\footnote{For simplicity, in the rest of this paper, we use the term "slot" to refer to "domain-slots"}

\subsection{Datasets}
\begin{savenotes}
\begin{table*}[]
\fontsize{8.5}{10.2}\selectfont
\centering
\begin{tabular}{@{}r|ccc|ccccc@{}}
\toprule
 &
  \textbf{DSTC2} &
  \textbf{MultiWOZ} &
  \textbf{SGD} &
  \textbf{SMCalFlow} &
  \textbf{ABCD} &
  \textbf{SIMMC2.0} &
  \textbf{BiToD} &
  \textbf{DSTC10} \\ \midrule
\textbf{Language(s)}         & en       & en, 7 lang.*     & en, ru, ar, id, sw & en      & en      & en          & en, zh           & en       \\
\textbf{Collection}          & H2M      & H2H              & M2M                & H2H     & H2H     & M2M         & M2M              & H2H      \\
\textbf{Modality}            & speech   & text             & text               & text    & text    & text, image & text             & speech$^{\ddagger}$   \\
\textbf{No. domains}         & 1        & 7                & 16                 & 4       & 30      & 1           & 5                & 3        \\
\textbf{No. dialogues}       & 1612     & 8438             & 16,142             & 41,517  & 8034    & 11,244      & 5787             & 107      \\
\textbf{No. turns}           & 23,354   & 115,424          & 329,964            & 155,923 & 177,407 & 117,236     & 115,638          & 2292     \\
\textbf{Avg. turns / dial. } & 14.5     & 13.7             & 20.4               & 8.2     & 22.1    & 10.4        & 19.9             & 12.6     \\
\textbf{Avg. tokens / turn}  & 8.5      & 13.8$^{\dagger}$ & 9.7$^{\dagger}$    & 10.2    & 9.2     & 12.8        & 12.2$^{\dagger}$ & 17.8     \\
\textbf{Dialogue state}      & slot-val & slot-val         & slot-val           & program & action  & slot-val    & slot-val         & slot-val \\
\textbf{Slots}               & 8        & 25               & 214                & -       & 231     & 12          & 68               & 18       \\
\textbf{Values}              & 212      & 4510             & 14,139             & -       & 12,047  & 64          & 8206             & 327      \\ \bottomrule
\end{tabular}
\caption{Characteristics of the main (left) and recent (right) datasets available for dialogue state tracking. *Machine translated. † Average for English. ‡ In the form of ASR hypotheses.}
\label{tab:datasets}
\end{table*}
\end{savenotes}

Many public datasets have been published to advance machine learning approaches for DST. The evolution of these datasets is marked by increasing dialogue complexity, especially with the advent of multidomain corpora.
We distinguish two main approaches for collecting dialogue datasets:
(i) Wizard-of-Oz (WOZ or H2H for \textit{human-to-human}) approaches where two humans (asynchronously) play the roles of user and agent according to a task description. This approach allows for natural and varied dialogues, but the subsequent annotation can be a source of errors.
(ii) Simulation-based approaches (M2M for \textit{machine-to-machine}) where two systems play the roles of user and agent and interact with each other to generate conversation templates that are then paraphrased by humans. The advantage of this method is that the annotations are obtained automatically. However, the complexity of the task and linguistic diversity are often limited because the dialogue is simulated.

Table \ref{tab:datasets} lists the main datasets as well as recent datasets relevant to the problems discussed in Section \ref{sec:defis}. 
In addition to the datasets mentioned, there are other less recent corpora listed in the survey of \citet{balaramanRecentNeuralMethods2021} that we do not include here for the sake of clarity. 
What follows is a description of the main datasets. 
Recent datasets, namely SMCalFlow\footnote{\url{https://microsoft.github.io/task_oriented_dialogue_as_dataflow_synthesis/}} \cite{andreasTaskOrientedDialogueDataflow2020}, ABCD\footnote{\url{https://github.com/asappresearch/abcd}} \cite{chenActionBasedConversationsDataset2021}, SIMMC 2.0\footnote{\url{https://github.com/facebookresearch/simmc2}} \cite{kotturSIMMCTaskorientedDialog2021}, BiToD\footnote{\url{https://github.com/HLTCHKUST/BiToD}} \cite{linBiToDBilingualMultiDomain2021} and DSTC10\footnote{\url{https://github.com/alexa/alexa-with-dstc10-track2-dataset}} \cite{kimHowRobustEvaluating2021a}, will be discussed in Section \ref{sec:defis}.

\paragraph{DSTC2\footnote{\url{https://github.com/matthen/dstc}} \cite{hendersonSecondDialogState2014}}
Early editions of the Dialogue State Tracking Challenge (DSTC) introduced the first shared datasets and evaluation metrics for DST and thus catalyzed research in this area. The corpus of the second edition remains a reference today. It includes dialogues between paid participants and various telephone dialogue systems (H2M collection). The user needs to find a restaurant by specifying constraints such as the type of cuisine and can request specific information such as the phone number.

\paragraph{MultiWOZ\footnote{\url{https://github.com/budzianowski/multiwoz}} \cite{budzianowskiMultiWOZLargeScalemultidomain2018}}
The first large-scale multidomain corpus and currently the main benchmark for DST. It contains dialogues between a tourist and a travel clerk that can span several domains.
A major problem related to the way the data was collected is the inconsistency and errors of annotation,  which was crowdsourced. Four more versions were later released to try and fix these errors. \cite{ericMultiWOZConsolidatedmultidomain2019, zangMultiWOZDialogueDataset2020, hanMultiWOZmultidomainTaskoriented2021, yeMultiWOZmultidomainTaskOriented2021}. 
Multilingual versions\footnote{Mandarin, Korean, Vietnamese, Hindi, French, Portuguese, and Thai.} were obtained by a process of machine translation followed by manual correction \cite{gunasekaraOverviewNinthDialog2020, zuoAllWOZMultilingualTaskOriented2021}.


\paragraph{SGD\footnote{\url{https://github.com/google-research-datasets/dstc8-schema-guided-dialogue}} \cite{rastogiScalablemultidomainConversational2020}} 
The Schema-Guided Dataset was created to elicit research on domain independence through the use of schemas. Schemas describe domains, slots, and intents in natural language and can be used to handle unseen domains. 
The test set includes unseen schemas to encourage model generalization. 
SGD-X is an extension designed to study model robustness to different schema wordings  \cite{leeSGDXBenchmarkRobust2021}. 

\subsection{Evaluation Metrics}

As there is a strict correspondence between dialogue history and dialogue state, DST models' performance is measured with accuracy metrics. Two metrics introduced by \citet{williamsDialogStateTracking2013} are commonly used for joint and individual evaluation.

\paragraph{Joint goal accuracy (JGA)}
Main metric referring to the set of user goals.
JGA indicates the performance of the model in correctly predicting the dialogue state at a given turn. It is equivalent to the proportion of turns where all predicted values for all slots exactly match the reference values.
A similar metric is the average goal accuracy which only evaluates predictions for slots present in the reference dialogue state \cite{rastogiSchemaGuidedDialogueState2020}.

\paragraph{Slot accuracy}
Unlike JGA, this metric evaluates the predicted value of each slot individually at each turn. It is computed as a macro-averaged accuracy: $SA = \frac{\sum_i^n acc_i}{n}$, where $n$ represents the number of slots.
For a more detailed evaluation, it can be decomposed according to slot type (e.g. requested slots, typically measured with the F1 score).

\paragraph{Alternative metrics}
Though these two metrics are widely used, they can be difficult to interpret. Slot accuracy tends to overestimate performance as most slots are not mentioned in a given turn while JGA can be too penalizing as a single wrong value prediction will result in wrong dialogue states for the rest of the conversation when accumulated through subsequent turns.
Two new metrics have been recently proposed to complement existing metrics and address these shortcomings: Relative slot accuracy ignores unmentioned slots and rewards the model for correct predictions \cite{kimMismatchMultiturnDialogue2022a}. Flexible goal accuracy is a generalized version of JGA which is more tolerant of errors that stem from an earlier turn \cite{deyFairEvaluationDialogue2022a}.

\subsection{Modern Approaches} \label{current}

One feature that categorizes DST models is the way they predict slot values. The prediction can be made either from a predefined set of values (\textit{fixed ontology}) or from an open set of values (\textit{open vocabulary}). What follows is a description of these approaches. A selection of recent models is presented in Table \ref{tab:models} based on this taxonomy.

\begin{table*}[h]
\fontsize{8.15}{9.78}\selectfont
\centering
\begin{tabular}{@{}lccccc@{}}
\toprule
\textbf{Model}                                         & \textbf{Decoder}                & \textbf{Context}                    & \textbf{Extra supervision} & \textbf{Ontology}   & \textbf{MWOZ2.1} \\ \midrule
\textbf{TRADE} \cite{wuTransferablemultidomainState2019}               & Generative & Full history          & -                         & \xmark       & 45.60    \\
\textbf{TOD-BERT} \cite{wuTODBERTPretrainedNatural2020}          & Classifier             & Full history          & Pretraining                         & \cmark       & 48.00   \\
\textbf{NADST}  \cite{leNonAutoregressiveDialogState2020}             & Generative               & Full history          & -                         & \xmark       & 49.04   \\
\textbf{DS-DST} \cite{zhangFindClassifyDual2020}          & Extract. + classif. & Previous turn              & -                         & Cat. slots & 51.21   \\
\textbf{SOM-DST}  \cite{kimEfficientDialogueState2020}          & Generative               & History + prev. state & -                         & \xmark       & 52.57   \\
\textbf{MinTL}  \cite{linMinTLMinimalistTransfer2020}          & Generative               & History + prev. state & Response generation                & \xmark       & 53.62   \\
\textbf{SST} \cite{chenSchemaGuidedmultidomainDialogue2020}                      & Classifier               & Prev. turn and state     & Schema graph         & \cmark       & 55.23   \\
\textbf{TripPy} \cite{heckTripPyTripleCopy2020} & Extractive & Full history & -                            & \xmark & 55.30  \\
\textbf{SimpleTOD} \cite{hosseini-aslSimpleLanguageModel2020} & Generative               & Full history          & TOD tasks                & \xmark       & 55.76   \\
\textbf{Seq2Seq-DU} \cite{fengSequencetoSequenceApproachDialogue2021}       & Generative               & Full history          & Schema                   & Cat. slots       & 56.10   \\
\textbf{SGP-DST}  \cite{leeDialogueStateTracking2021}                      & Generative               & Full history          & Schema                   & Cat. slots & 56.66   \\
\textbf{SOLOIST} \cite{pengSOLOISTBuildingTask2021}                         & Generative               & Full history          & Pretraining                & \xmark       & 56.85   \\
\textbf{PPTOD} \cite{suMultiTaskPreTrainingPlugandPlay2022} & Generative               & Full history          & Pretrain + TOD tasks                & \xmark       & 57.45   \\
\textbf{D3ST} \cite{zhaoDescriptionDrivenTaskOrientedDialog2022}               & Generative               & Full history          & Schema                   & \xmark       & 57.80    \\ 
\textbf{TripPy + SCoRe} \cite{yuSCoRePreTrainingContext2021}     & Extractive & Full history & Pretraining & \xmark & 60.48 \\
\textbf{TripPy + CoCo} \cite{liCoCoControllableCounterfactuals2021}      & Extractive & Full history & Data augmentation & \xmark & 60.53 \\
\textbf{TripPy + SaCLog} \cite{daiPreviewAttendReview2021}     & Extractive & Full history & Curriculum learning & \xmark & 60.61 \\
\textbf{DiCoS-DST} \cite{guoGranularityMultiPerspectiveDialogue2022}               & Extract. + classif.               & Relevant turns          & Schema graph                   & \cmark       & 61.02    \\ 
\bottomrule
\end{tabular}
\caption{Characteristics of recent DST models and performance in terms of joint goal accuracy on MultiWOZ 2.1 \cite{ericMultiWOZConsolidatedmultidomain2019}. "Ontology" denotes access to a predefined set of values for each slot.}
\label{tab:models}
\end{table*}

\paragraph{Fixed ontology}
Following the discriminative approaches that preceded them, traditional neural approaches are based on a fixed ontology and treat DST as a multiclass classification problem \cite{hendersonWordBasedDialogState2014a, mrksicNeuralBeliefTracker2017}. Predictions for a given slot are estimated by a probability distribution over a predefined set of values, restricting the prediction field to a closed vocabulary and thus simplifying the task considerably.
The performance of this approach is therefore relatively high \cite{chenSchemaGuidedmultidomainDialogue2020}, however, its cost is proportional to the size of the vocabulary as all potential values have to be evaluated. In practice, the number of values can be large and a predefined ontology is rarely available.

\paragraph{Open vocabulary}
To overcome these limitations, approaches to predict on an open set of values have been proposed.
The first method consists in extracting values directly from the dialogue history, e.g. by formulating DST as a reading comprehension task \cite{gaoDialogStateTracking2019}.
This method depends solely on the dialogue context to extract value spans, however, slot values can be implicit or have different wordings (e.g. the value "expensive" may be expressed as "high-end").
An alternative is to generate slot values using an encoder-decoder architecture.
For instance, TRADE uses a copy mechanism to generate a value for each slot based on a representation of the dialogue history \cite{wuTransferablemultidomainState2019}. A common current approach is to decode the dialogue state using a pretrained autoregressive language model \cite{hosseini-aslSimpleLanguageModel2020}.


\paragraph{Hybrid methods} 
A trade-off seems to exist between the level of value independence in a model and DST performance. Some works have sought to combine fixed ontology approaches with open vocabulary prediction to benefit from the advantages of both methods.
This approach is based on the distinction between categorical slots for which a set of values is predefined, and non-categorical slots with an open set of values \cite{goelHySTHybridApproach2019a, zhangFindClassifyDual2020, heckTripPyTripleCopy2020}.

\section{Recent Advances}  \label{sec:recent}
In recent years, several important problems have been addressed, notably through the use of pretrained language models (PLMs) trained on large amounts of unannotated text.
This section summarizes the advances in 2021-2022. 

\subsection{Modeling Slot Relationships}
The methods mentioned so far treat slots individually without taking into account their relations. However, slots are not conditionally independent, for instance, slot values can be correlated, e.g. hotel stars and price range.
An alternative is to explicitly consider these relations using self-attention \cite{yeSlotSelfAttentiveDialogue2021}.
In a similar vein, \citet{linKnowledgeAwareGraphEnhancedGPT22021} adopt a hybrid architecture to enable sequential value prediction from a GPT-2 model while modeling the relationships between slots and values with a graph attention network (GAT).

Rather than learning these relationships automatically, another line of work uses the knowledge available from the domain ontology, for example by taking advantage of its hierarchical structure \cite{liGenerationExtractionCombined2021}. 
These relationships can also be represented in a graph with slots and domains as nodes.
\citet{chenSchemaGuidedmultidomainDialogue2020} first build a schema graph based on the ontology and then use a GAT to merge information from the dialogue history and the schema graph. 
\citet{fengDynamicSchemaGraph2022} extend this approach by dynamically updating slot relations in the schema graph based on the dialogue context. \citet{guoGranularityMultiPerspectiveDialogue2022} incorporate both dialogue turns and slot-value pairs as nodes to consider relevant turns only and solve implicit mentions.



\subsection{Adapting PLMs to Dialogues}
Though now commonly used for DST, existing PLMs are pretrained on free-form text using language modeling objectives. Their ability to model dialogue context and multi-turn dynamics is therefore limited. It has been shown that adapting a PLM to the target domain or task by continuing self-supervised learning can lead to performance gains \cite{gururanganDonStopPretraining2020}. 
This method has been applied to TOD systems and DST. 

There are two underlying questions with this approach: the selection of adaptation data and the formulation of self-supervised training objectives to learn better dialogue representations for the downstream task.
\citet{wuTODBERTPretrainedNatural2020} gather nine TOD corpora and continue BERT's pretraining with masked language modeling and next response selection. The obtained model TOD-BERT provides an improvement over a standard BERT model on several TOD tasks including DST.
With a similar setup, \citet{zhuWhenDoesFurther2021} contrast these results and find that such adaptation is most beneficial when little annotated data is available.
Based on TOD-BERT, \citet{hungDSTODEfficientDomain2022} show that it is advantageous not only to adapt a PLM to dialogues but also to the target domain. To do so, they use conversational data from Reddit filtered to contain terms specific to the target domain. 
Finally, \citet{yuSCoRePreTrainingContext2021} introduce two objective functions designed to inject inductive biases into a PLM in order to jointly represent dynamic dialogue utterances and ontology structure. They evaluate their method on conversational semantic parsing tasks including DST.

\subsection{Mitigating Annotated Data Scarcity}
The lack of annotated data hinders the development of efficient and robust DST models. However, the data collection process is costly and time-consuming. 
One approach to address this problem is to train a model on resource-rich domains and apply it to an unseen domain with little or no annotated data (cross-domain transfer; \citealp{wuTransferablemultidomainState2019}).
\citet{dingliwalFewShotDialogue2021} adopt meta-learning and use the source domains to meta-learn the model's parameters and initialize fine-tuning for the target domain.
Works around schema-based datasets \cite{rastogiSchemaGuidedDialogueState2020} use slot descriptions to handle unseen domains and slots \cite{linLeveragingSlotDescriptions2021, zhaoDescriptionDrivenTaskOrientedDialog2022}.
A drawback of these approaches is that they rely on the similarity between the unseen domain and the initial fine-tuning domains.

Another set of approaches tries to exploit external knowledge from other tasks with more abundant resources.
\citet{hudecekDiscoveringDialogueSlots2021} use FrameNet semantic analysis as weak supervision to identify potential slots. 
\citet{gaoMachineReadingComprehension2020, liZeroshotGeneralizationDialog2021, linZeroShotDialogueState2021} propose different methods to pretrain a model on reading comprehension data before applying it to DST. Similarly, \cite{shinDialogueSummariesDialogue2022a} reformulate DST as a dialogue summarization task based on templates and leverage external annotated data.

Note that the PLM adaptation approaches seen above allow for more efficient learning when little data is available and are also a potential solution to the data scarcity problem. Along this line, \citet{miSelftrainingImprovesPretraining2021} present a self-learning method complementary to TOD-BERT for few-shot DST. 


\subsection{Prompting Generative Models to Address Unseen Domains}
\begin{figure}[h]
\centering
\includegraphics[width=0.51\textwidth]{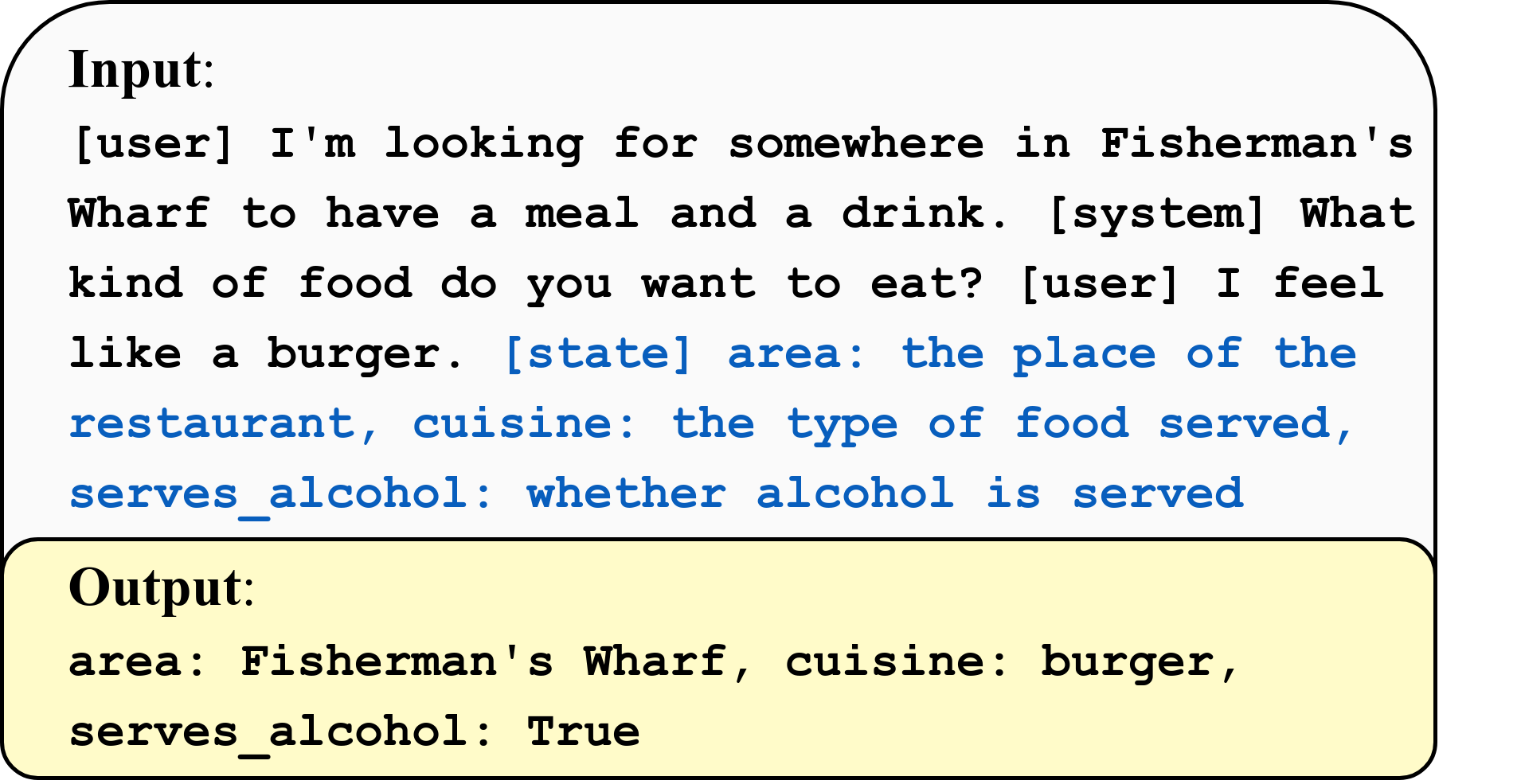}
\caption{An example of text-to-text DST from SGD \cite{rastogiScalablemultidomainConversational2020}. Including \textcolor{Mycolor3}{slot descriptions} as a prompt makes it possible to address unseen domains.}
\label{fig:seq2seq}
\end{figure}

A recent paradigm tackles all text-based language tasks with a single model by converting them into a text-to-text format \cite{raffelExploringLimitsTransfer2020}. This approach relies on textual instructions called prompts that are prepended to the input to condition the model to perform a task. It has been successfully applied to DST, not only closing the performance gap between classification and generation methods but also opening opportunities to address unseen domains using schemas (cfr. Section \ref{sec:dst}). In addition to slot and domain names, \citet{leeDialogueStateTracking2021} include slot descriptions in a prompt to independently generate values for each slot. They also add possible values for categorical slots. 
\citet{zhaoEffectiveSequencetoSequenceDialogue2021a, zhaoDescriptionDrivenTaskOrientedDialog2022} expand on this approach by generating the entire dialogue state sequentially, as illustrated in Figure \ref{fig:seq2seq}.
In an analysis of prompt formulation,  \citet{caoComparativeStudySchemaGuided2021} find that a question format yields better results for prompt-based DST.

Hand-crafting textual prompts may lead to sub-optimal conditioning of a PLM. 
Instead of using discrete tokens, we can optimize these instructions as continuous embeddings, a method called prompt tuning \cite{lesterPowerScaleParameterEfficient2021}, which has been applied to DST for continual learning, adding new domains through time \cite{zhuContinualPromptTuning2022a}.  
Alternatively, \citet{yangPromptLearningFewShot2022} reverse description-based prompts and formulate a prompt based on values extracted from the utterance to generate their respective slot. They argue that slots that appear in a small corpus do not represent all potential requirements whereas values are often explicitly mentioned.
Rather than using descriptions that indirectly convey the semantics of a schema, others have sought to prompt a model with instructions, i.e. in-context learning, in which the prompt consists of a few example input-output pairs \cite{huInContextLearningFewShot2022, guptaShowDonTell2022}.



\subsection{Going Beyond English}
Until recently, most works around DST were confined to English due to the lack of data in other languages, preventing the creation of truly multilingual models.
In recent years, several works have addressed this issue. 
A DSTC9 track studied cross-lingual DST. For this purpose, a Chinese version of MultiWOZ and an English version of CrossWOZ were obtained by a process of machine translation followed by manual correction \cite{gunasekaraOverviewNinthDialog2020}. 
\citet{zuoAllWOZMultilingualTaskOriented2021} used the same method to translate MultiWOZ in seven languages.

A problem with machine translations is that they lack naturalness and are not localized. 
Two Chinese datasets were obtained by H2H collection: CrossWOZ and RiSAWOZ \cite{zhuCrossWOZLargeScaleChinese2020, quanRiSAWOZLargeScaleMultiDomain2020}, however, this type of collection is expensive. 
The M2M approach makes it possible to obtain adequate multilingual corpora by adapting conversation templates according to the target language.
\citet{linBiToDBilingualMultiDomain2021} took advantage of this method to create BiToD, a bilingual English-Chinese corpus.
\citet{majewskaCrossLingualDialogueDataset2022} used schemas from SGD to create conversation templates that were then adapted into Russian, Arabic, Indonesian and Swahili.
Similarly, \citet{dingGlobalWoZGlobalizingMultiWoZ2022} translated templates from the MultiWOZ test set in Chinese, Spanish and Indonesian and localized them.
Lastly, to overcome the lack of multilingual data, \citet{mogheCrosslingualIntermediateFinetuning2021} leverage parallel and conversational movie subtitle data to pretrain a cross-lingual DST model.



\section{Challenges and Future Directions} \label{sec:defis}

Despite recent advances, there are still many challenges in developing DST models that can accurately capture user needs dynamically and in a variety of scenarios. 
There are many interesting paths, this section articulates the needs around three axes: generalization, robustness and relevance of models.

\subsection{Generalizability}
TOD systems are intended to be deployed in dynamic environments that may involve different settings. 
In practice, the application domains of these systems are numerous and varied (e.g. customer service in telecommunications, banking, technical support, etc.), which makes manual annotation of corpora for each domain difficult or impossible. This reduces the effectiveness of traditional supervised learning and is one of the reasons why most systems in production are rule-based. 

Learning with little or no new annotated data offers an alternative 
to take advantage of the capabilities of neural networks
to guarantee the flexibility of the systems. 
The importance of this aspect is reflected by the numerous recent works that address this problem, as presented in Section \ref{sec:recent}. 
Although significant progress has been made, these approaches remain limited. Models that rely on existing DST resources are unable to handle domains whose distribution deviates from that of the training data \cite{dingliwalFewShotDialogue2021}. Models that use external knowledge offer a more generic approach but achieve relatively poor performance \cite{linZeroShotDialogueState2021}.
Learning generalizable models thus remains an open problem. 
An interesting avenue is continual learning, which allows new skills to be added to a system over time after deployment. Without retraining with all the data, the model must be able to accumulate knowledge \cite{madottoContinualLearningTaskOriented2021, liuDomainLifelongLearningDialogue2021, zhuContinualPromptTuning2022a}.

In a real scenario, entities that are not observed during training are bound to appear. A DST model must be able to extrapolate from similar entities that have been seen.
However, MultiWOZ has been shown to exhibit an entity bias, i.e. the slot values distribution is unbalanced. When a generative DST model was evaluated on a new test set with unseen entities, its performance dropped sharply, hinting at severe memorization \cite{qianAnnotationInconsistencyEntity2021}. 
Similarly, by its constrained nature, a TOD system may be perturbed by out-of-domain utterances. It is desirable to be able to recognize such utterances in order to provide an appropriate response.
The ability of a model to generalize to new scenarios is therefore related to its robustness. 

\subsection{Robustness}
Since users may express similar requirements in different ways, a DST model must be able to interpret different formulations of a request in a consistent manner. In other words, it must be robust to variations in the input. Analytical work has shown that models' performance drops when they are confronted with realistic examples that deviate from the test set distribution \cite{huangCounterfactualMattersIntrinsic2021, liCoCoControllableCounterfactuals2021}. Recent work has tried to address this issue through regularization techniques \cite{heckRobustDialogueState2022}.
Related to this, an understudied aspect is dealing with utterances that deviate from the norm in the case of a written dialogue system.

A DST model must be able to take into account all the history and adjust its predictions 
of the dialogue state using all available information.
Many works have found that performance degrades rapidly as the dialogue length increases. Another critical aspect is therefore efficiently processing long dialogues \cite{zhangImprovingLongerrangeDialogue2021}.
The dialogue state condenses important information, but correcting an error made in an earlier turn may be difficult. To overcome this error propagation issue, \citet{tianAmendableGenerationDialogue2021} use a two-pass dialogue state generation to correct potential errors, while
\citet{manotumruksaImprovingDialogueState2021} propose a turn-based objective function to penalize the model for incorrect prediction in early turns.
Despite this need, \citet{jakobovitsWhatDidYou2022} have shown that popular DST datasets are not conversational: most utterances can be parsed in isolation.
Some works simulate longer and more realistic dialogues by inserting chit-chat into TODs \cite{kotturDialogStitchSyntheticDeeper2021, sunAddingChitChatEnhance2021}.


Another point that has not been studied much in recent approaches is robustness to speech inputs \cite{faruquiRevisitingBoundaryASR2021a}. For spoken dialogue systems, new challenges arise such as ASR errors or verbal disfluencies.
Early editions of DSTC provided spoken corpora that included a transcript and ASR hypotheses. 
Since then, DST datasets have been primarily text-based.
A DSTC10 track considered this aspect again and proposed a DST task with validation and test sets containing ASR hypotheses \cite{kimHowRobustEvaluating2021a}. 

Learning robust models requires diverse datasets that represent real-world challenges.
In this sense, several evaluation benchmarks have been published to study TOD systems' robustness \cite{leeSGDXBenchmarkRobust2021, pengRADDLEEvaluationBenchmark2021, choCheckDSTMeasuringRealWorld2021}.
Data augmentation is a potential solution to the lack of variety in datasets \cite{campagnaZeroShotTransferLearning2020, liCoCoControllableCounterfactuals2021, aksu-etal-2022-n}, especially for simulating ASR errors \cite{wangDataAugmentationTraining2020, tianTODDABoostingRobustness2021}.

\subsection{Relevance}
As we have seen, existing datasets do not really reflect real-world conditions resulting in a rather artificial task. It is important to keep a holistic view of the development of DST models to ensure the relevance of their application in a dialogue system.

Since the first TOD systems, the dialogue state has been considered as a form to be filled in as slot-value pairs. 
This fixed representation is suitable for simple tasks like flight booking but is limited in domains with rich relational structures and a variable number of entities.
Indeed, composition is not possible (e.g. "itinerary for my next meeting") and knowledge is not directly shared between slots.
Section \ref{sec:recent} presented approaches that attempt to address the latter point by using graphs for dialogue state representation. 
To promote work on more realistic scenarios, some have proposed richer representations with an associated corpus. 
\citet{andreasTaskOrientedDialogueDataflow2020} encode the dialogue state as a data-flow graph and introduce the SMCallFlow corpus.
\citet{chengConversationalSemanticParsing2020} propose a tree structure along with the TreeDST corpus. ThingTalk is another alternative representation of the dialogue state which was successfully applied to MultiWOZ \cite{lamThingTalkExtensibleExecutable2022, campagnaFewShotSemanticParser2022}. 
In the ABCD corpus, \citet{chenActionBasedConversationsDataset2021} adopt a representation of the procedures that a customer service employee must follow in accordance with company policies. 

These approaches still rely on specific database schemas and are limited to one modality. For more capable virtual agents, we can extend the scope of dialogues to a multimodal world. Emerging efforts on multimodal DST seek to track the information of visual objects based on a multimodal context.
With SIMMC 2.0, \citet{kotturSIMMCTaskorientedDialog2021} introduced a multimodal TOD dataset based on virtual reality rendered shopping scenes.
Similarly, \citet{le-etal-2022-multimodal} proposed a synthetic dataset of dialogues consisting of multiple question-answer pairs about a grounding video input. The questions are based on 3D objects. In both these datasets, the system has to track visual objects in the dialogue state in the form of slot-value pairs.
For a broader background on multimodal conversational AI, we refer the reader to \citet{sundar-heck-2022-multimodal}'s survey. 

Dialogue is dynamic: in a real scenario, an erroneous prediction of the dialogue state would have deviated the course of the conversation from the reference dialogue.
However, most studies evaluate models in isolation, assuming that it is always possible to assemble a set of well-performing components to build a good TOD system.
The overall performance of a system is rarely taken into account as evaluating the system as a whole is complicated and requires human evaluation. 
Moreover, it can be difficult to identify which component of the system is problematic and needs to be improved.
Despite these hurdles, it is important to consider the impact that DST can have on the dialogue system as a whole. 
\citet{takanobuYourGoalOrientedDialog2020a} conducted automatic and human evaluations of dialogue systems with a wide variety of configurations and settings on MultiWOZ. They found a drop in task success rate using DST rather than NLU followed by a rule-based dialogue state update. They explain this result by the fact that NLU extracts the user's intentions in addition to the slot-value pairs. 
Another work showed how uncertainty estimates in belief tracking can lead to a more robust downstream policy \cite{vanniekerkUncertaintyMeasuresNeural2021}.
These studies are rare in their kind and call for more similar work. 

\section{Conclusion}
Dialogue state tracking is a crucial component of a conversational agent to identify the user's needs at each turn of the conversation.  
A growing body of work is addressing this task and we have outlined the latest developments.
After giving an overview of the task and the different datasets available, we have categorized modern neural approaches according to the inference of the dialogue state. 
Despite encouraging results on benchmarks such as MultiWOZ, these systems lack flexibility and robustness, which are critical skills for a dialogue system.
In recent years, many works have sought to address these limitations and we have summarized the advances.
However, there are still significant challenges to be addressed in the future.
There are many interesting avenues and we have proposed three key features of DST models to guide future research: generalizability, robustness and relevance.

\section*{Acknowledgements}
We thank the anonymous reviewers for their valuable feedback and suggestions for further improvement.

\bibliography{anthology,custom}
\bibliographystyle{acl_natbib}

\appendix



\end{document}